\documentclass[11pt]{article}
\usepackage[margin=0.5in]{geometry}
\usepackage{amsmath,graphicx,hyperref}
\usepackage{graphicx}
\usepackage{xcolor}
\usepackage{booktabs}
\usepackage{multirow}
\usepackage{amsmath,amsfonts,amssymb}
\usepackage{array}
\usepackage{makecell}
\usepackage[table]{xcolor}
\usepackage{bbm}
\usepackage[most]{tcolorbox}
\usepackage{listings}
\usepackage{algorithm}
\usepackage{algorithmic}
\usepackage{dsfont}
\usepackage{pifont}
\usepackage{verbatim}
\usepackage{enumitem}
\usepackage[most]{tcolorbox}
\definecolor{darkerblue}{RGB}{110, 145, 195} 
\definecolor{darkerred}{RGB}{180, 125, 120} 
\definecolor{darkeryellow}{RGB}{205, 180, 90} 
\definecolor{darkerorange}{RGB}{215, 160, 100} 
\definecolor{darkerpurple}{RGB}{130, 100, 160}

\definecolor{green1}{HTML}{A8E6CF}
\definecolor{green2}{HTML}{D4F5C7}
\definecolor{red1}{HTML}{E8A1A1}
\definecolor{red2}{HTML}{F5BDBD}
\definecolor{yellow1}{HTML}{FFF7C2} 
\definecolor{orange1}{HTML}{FFE4B2}
\definecolor{blue1}{HTML}{C9E4F6}
\definecolor{purple1}{HTML}{D9C9EB}
\definecolor{pink1}{HTML}{F4C6D7}
\definecolor{cgreen}{HTML}{228B22}
\newcommand{\cmark}{\textcolor{cgreen}{\ding{51}}}
\newcommand{\xmark}{\textcolor{red}{\ding{55}}}
\definecolor{mygreen}{RGB}{64,140,120} 
\newtcolorbox{highlightbox}{
  colback=green,
  colframe=mygreen,
  boxrule=1.2pt,
  arc=3mm,     
  left=6mm, right=6mm, top=4mm, bottom=4mm,
  enhanced,
}
\usepackage[table]{xcolor}
\usepackage[table]{xcolor}
\newcommand{\posval}{\textcolor{green!60!black}{+}}
\newcommand{\negval}{\textcolor{red!70!black}{-}}
\newcommand{\posmain}[1]{\textcolor{green!60!black}{#1}}
\newcommand{\negmain}[1]{\textcolor{red!70!black}{#1}}
\usepackage{comment}
\usepackage{xcolor}
\definecolor{conservative}{RGB}{30,80,180}
\definecolor{balanced}{RGB}{46,125,50}
\definecolor{aggressive}{RGB}{140,70,20}
\definecolor{poor}{RGB}{94,53,177}
\definecolor{zcolor}{RGB}{170,90,85}
\usepackage{authblk}
\title{Conversational Speech Reveals Structural Robustness Failures\\ in SpeechLLM Backbones}

\author[1]{Maria Teleki}
\author[1]{Sai Tejas Janjur}
\author[1]{Haoran Liu}
\author[1]{Oliver Grabner}
\author[1]{Ketan Verma}
\author[1]{Thomas Docog}
\author[1]{Xiangjue Dong}
\author[1]{Lingfeng Shi}
\author[1]{Cong Wang}
\author[1]{Stephanie Birkelbach}
\author[1]{Jason Kim}
\author[1]{Yin Zhang}
\author[2]{Éva Székely}
\author[1]{James Caverlee}

\affil[1]{Texas A\&M University, USA}
\affil[2]{KTH Royal Institute of Technology, Sweden}

\date{}

\begin{document}

\maketitle

\vspace{-2em}


\footnotetext{Under review at INTERSPEECH '26.}

\begin{abstract}
    LLMs serve as the backbone in SpeechLLMs, yet their behavior on spontaneous conversational input remains poorly understood. Conversational speech contains pervasive disfluencies -- interjections, edits, and parentheticals -- that are rare in the written corpora used for pre-training. Because gold disfluency removal is a deletion-only task, it serves as a controlled probe to determine whether a model performs faithful structural repair or biased reinterpretation. Using the DRES evaluation framework, we evaluate proprietary and open-source LLMs across architectures and scales. We show that model performance clusters into stable precision-recall regimes reflecting distinct ``editing policies.'' Notably, reasoning models systematically over-delete fluent content, revealing a bias toward semantic abstraction over structural fidelity. While fine-tuning achieves SOTA results, it harms generalization. Our findings demonstrate that robustness to speech is shaped by specific training objectives.
\end{abstract}

\section{Introduction}
\label{Introduction}

As SpeechLLMs become central to voice assistants, meeting transcription, and multimodal conversational systems, a key assumption has emerged: increasing model scale and reasoning capability improves robustness to real-world speech \cite{ma2025audiocotexploringchainofthoughtreasoning, radford2022robustspeechrecognitionlargescale, hu2024wavllmrobustadaptivespeech, fan2026incentivizing}. We show that this assumption is incomplete. Conversational speech reveals structural failure patterns in current LLM backbones that are not clearly captured by standard semantic benchmarks or aggregate end-to-end metrics.

Spontaneous conversational speech contains pervasive disfluencies -- interjections (\textit{uh}, \textit{um}), repetitions, false starts, and parentheticals (\textit{you know}, \textit{I mean}) -- that are intrinsic to incremental human production \cite{alter2013benefits, Shriberg.1994, Shriberg_1996, bortfeld2001disfluency, Clark_Tree_2002, meteer1995dysfluency, diachek2023effect}. Consider the utterance:

\begin{quote}
    \textit{``\underline{I uh I mean} the other driver \underline{was} — was going through the red light when the crash happened.''}
\end{quote}

A fluent representation of this utterance is:

\begin{quote}
    \textit{``The other driver was going through the red light when the crash happened.''}
\end{quote}

The transformation required here is subtractive: disfluent spans are removed while the fluent content is otherwise preserved. In other words, the fluent transcript corresponds to a monotonic subsequence of the original input. This constraint makes disfluency removal a tightly controlled transformation: there is no paraphrastic freedom, and additional deletion or rewriting constitutes a structural error.

Yet large generative models are optimized for abstraction, compression, and semantic reinterpretation. These objectives conflict with deletion-constrained repair, which requires strict preservation of the original token sequence while removing only disfluent spans. As a result, models may rewrite or reinterpret conversational structure rather than faithfully repairing it. However, conversational disfluencies carry important paralinguistic signals -- for example, filled pauses can convey speaker uncertainty or cognitive state \cite{brennan1995feeling, kirkland2022s, dinkar2023fillersspokenlanguageunderstanding}. Misinterpreting these structures can therefore have downstream consequences in high-stakes settings, including linguistic forensics and judicial decision-making \cite{torstenston2009discourse, harrington2021style, schiel2015disfluencies, larson2004disfluencies, hernandez2013disfluency}, medical documentation and decision-making \cite{10.1145/3630106.3658996}, and social reasoning tasks \cite{loy2019real} including personality assessment \cite{wester2015artificial} and deception detection \cite{de2024don, loy2016lying}. We therefore ask:

\begin{quote}
    \textbf{\textit{Are robustness failures in SpeechLLMs driven by limitations in how their LLM backbones handle conversational speech?}}
\end{quote}

\noindent To investigate this question, we introduce a controlled probe: conversational disfluency removal. Given gold annotations, the correct output is uniquely defined by a deletion-only transformation. The resulting fluent transcript must preserve all fluent tokens from the input while removing only annotated disfluencies. Under this constraint, robustness reduces to token-level agreement with a gold deletion mask. Any over-deletion or under-deletion directly reveals structural editing errors.

We operationalize this idea through the \textbf{Disfluency Removal Evaluation Suite (DRES)}, a structural evaluation framework designed to isolate language-level editing behavior in SpeechLLM backbones. Unlike end-to-end speech benchmarks, which conflate acoustic transcription errors with language-level editing decisions, DRES evaluates models using fixed gold conversational transcripts. This factorization allows \textbf{structural editing behavior} to be measured independently of acoustic suppression effects.

Using DRES, we evaluate a diverse set of proprietary and open-source LLM backbones spanning architectures, parameter scales, prompting regimes, and reasoning variants (\autoref{tab:model-comparison}). Our analysis reveals that models exhibit stable editing policies characterized by distinct precision–recall trade-offs in deletion behavior. Notably, reasoning-oriented models tend to over-delete fluent content, reflecting a bias toward semantic abstraction rather than structural fidelity. While lightweight fine-tuning can substantially improve structural repair performance, it introduces measurable degradation on unrelated reasoning and knowledge benchmarks.

\begin{figure*}
    \centering
    \includegraphics[width=0.99\linewidth]{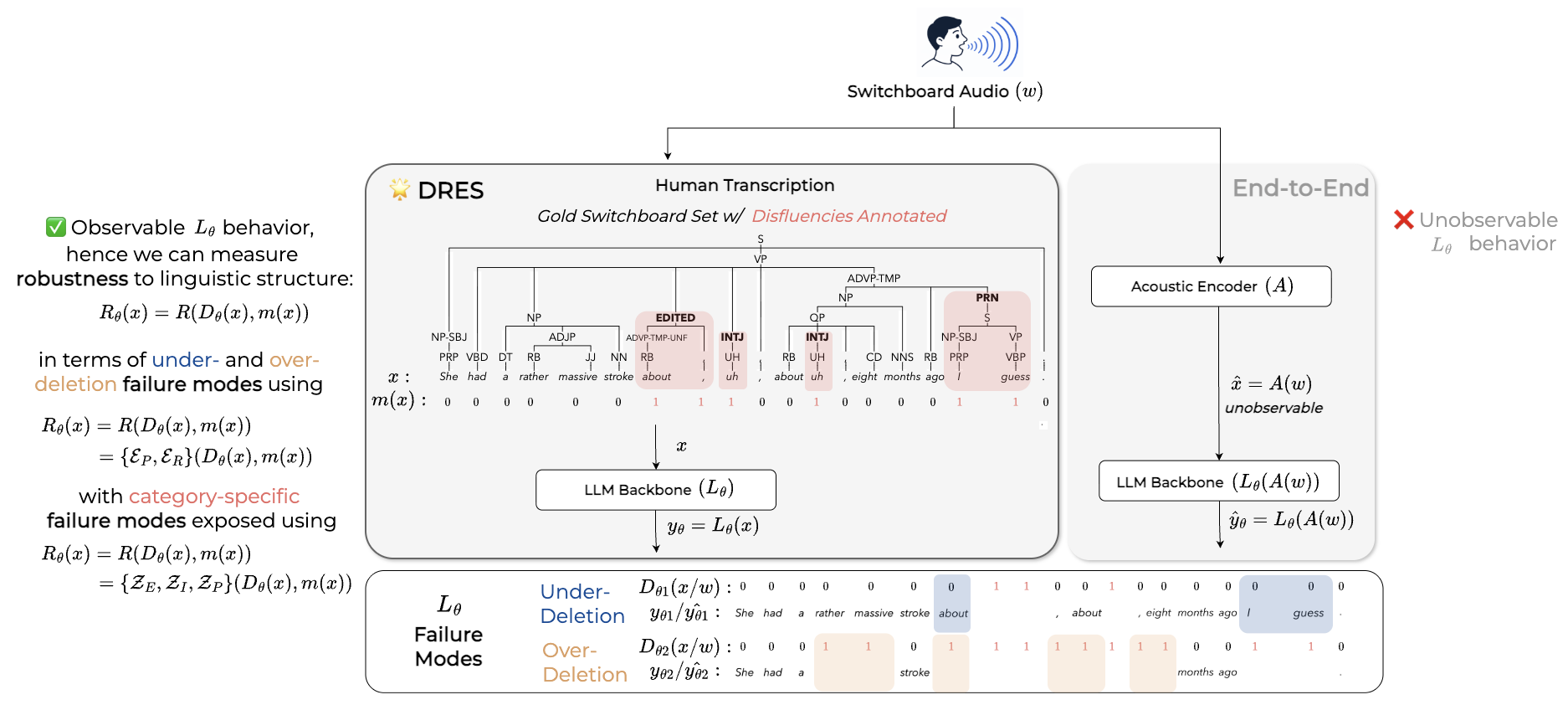}
    \caption{\textbf{Overview of DRES, a factorized structural evaluation framework for SpeechLLM backbones:} Gold Switchboard transcripts with annotated disfluencies are provided directly to the LLM backbone to isolate language-level deletion behavior ($D_{\theta}(x)$) from acoustic suppression effects (unobservable $A(w)$) in end-to-end SpeechLLMs. Structural robustness is computed by comparing the model-implied deletion mask to the gold mask, decomposing errors into (i) \textcolor{aggressive}{over-deletion} and \textcolor{conservative}{under-deletion} failure modes in the precision–recall space (\autoref{fig:main}), and (ii) \textcolor{zcolor}{category-specific} failure modes (\autoref{fig:clustering}).}
    \label{fig:system}
\end{figure*}

\smallskip
\noindent We contribute:

\begin{itemize}
\item \textbf{DRES: a structural evaluation framework for SpeechLLM backbones (\autoref{fig:system}, \S \ref{Robustness to Conversational Structure}).}
We introduce a factorized evaluation protocol that isolates language-level editing behavior by providing models with gold conversational transcripts and enforcing deletion-only constraints. We will open-source the code for DRES.

\item \textbf{A structural definition of conversational robustness (\S \ref{Robustness and Editing Policy Definitions}).}
We formalize robustness as deletion-constrained repair and measure it through token-level agreement with gold deletion masks, enabling direct analysis of over-deletion and under-deletion errors.

\item \textbf{Empirical identification of editing policies in LLM backbones (\autoref{fig:main}, \S \ref{Robustness to Conversational Structure}, \S \ref{Empirical Evidence of Policy-Level Robustness Patterns}).}
Across diverse proprietary and open-source models, we show that LLMs cluster into stable precision–recall regimes corresponding to \textcolor{conservative}{\textbf{under-deletion}}, \textcolor{aggressive}{\textbf{over-deletion}}, \textcolor{balanced}{\textbf{balanced}}, and \textcolor{poor}{\textbf{poor}} editing policies shaped by training objectives.

\item \textbf{Evidence of a robustness--generalization trade-off under adaptation (\S \ref{Finding 5}).}
Fine-tuning significantly improves structural fidelity on disfluency removal but degrades performance on reasoning and knowledge benchmarks, indicating a specialization cost.

\item \textbf{Practical deployment recommendations (\S \ref{Practical Recommendations}).} We translate these empirical findings into practical recommendations, including transcript segmentation for stability, editing-policy-aware model selection, and monitoring generalization degradation under fine-tuning.
\end{itemize}

\noindent Conversational speech therefore provides a controlled stress test for structural alignment in language models. As SpeechLLMs become increasingly integrated into real-world systems, evaluating robustness requires not only measuring semantic accuracy but also auditing how faithfully models preserve the structure of conversational language.

\section{Related Work}
\label{Related Work}

\subsection{SpeechLLMs and End-to-End Evaluation}
\label{SpeechLLMs and End-to-End Evaluation}

Recent SpeechLLM architectures combine acoustic encoders with pretrained LLM backbones to produce end-to-end systems mapping speech to text outputs \cite{yang-etal-2025-large-language, cui2025recent, arora2025landscape, Peng_2026}. Benchmarks such as VocalBench-DF \cite{liu2025vocalbenchdfbenchmarkevaluatingspeech}, WildSpeech-Bench \cite{zhang2025wildspeechbenchbenchmarkingendtoendspeechllms}, and VoxEval \cite{cui2025voxevalbenchmarkingknowledgeunderstanding}  evaluate system-level robustness to disfluent speech under realistic acoustic conditions. These efforts provide important insights into deployment behavior.

However, end-to-end evaluation entangles acoustic transcription and language-level editing policies. Modern ASR systems are known to under-transcribe certain disfluencies, especially short interjections, which can influence downstream task performance \cite{mujtaba2024lost, teleki24_interspeech, retkowski2025summarizingspeechcomprehensivesurvey}. Further, ASR systems are known to error specifically on disfluent speech, with harmful hallucinations \cite{ koeneckeperspective, 10.1145/3630106.3658996}.  As a result, improvements in system-level robustness may reflect acoustic suppression or expansion, language-level editing behavior, or their interaction.

DRES complements end-to-end benchmarks by operating on fixed gold transcripts and enforcing deletion-only evaluation, \textit{isolating how backbone LLMs understand highly disfluent conversational speech}. By removing acoustic variability, it isolates backbone editing behavior and enables direct measurement of structural fidelity in conversational repair.

\subsection{Disfluency Detection and Removal}
\label{Disfluency Detection and Removal}

Research on disfluency has largely focused on detection within conversational speech, particularly using the Switchboard corpus and its treebank annotations. Early approaches modeled repairs using noisy-channel and parsing-based methods that treated EDITED regions as structured syntactic objects \cite{Charniak.2001, johnson-charniak-2004-tag}. Subsequent work reframed disfluency detection as a sequence labeling problem, employing BiLSTM and Semi-CRF \cite{zayats2016disfluency} and EGBC \cite{bach2019noisy} and more. Recently, pretrained contextual encoders have been used to identify disfluent spans \cite{jamshid-lou-johnson-2020-improving}. The primary metrics for disfluency removal are word-level precision ($\mathcal{E}_P$), recall ($\mathcal{E}_R$), and F1 scores ($\mathcal{E}_F$). Recent work introduces a method to perform alignment between generative model outputs and gold parse trees, enabling this scoring for backbone LLMs \cite{teleki25_zscores}. Previous systems \cite{Charniak.2001, johnson-charniak-2004-tag, zayats2016disfluency, bach2019noisy, jamshid-lou-johnson-2020-improving} substantially improved token- and span-level F1, particularly for \textcolor{zcolor}{\textbf{INTJ}} and \textcolor{zcolor}{\textbf{PRN}} structures. A new standardized metric, $\mathcal{Z}$-Scores, measures performance on these structures at the span level \cite{teleki25_zscores}.

\subsection{Robustness in Large Language Models}
\label{Robustness in Large Language Models}

Robustness in large language models has largely been studied through adversarial perturbations, distribution shift, and reasoning benchmarks \cite{xu2024an, zhang2025evaluating, kung2023active, agrawal2025enhancingllmrobustnessperturbed}. These evaluations test whether models maintain semantic consistency under lexical variation, domain transfer, or multi-step inference stress tests. Performance is typically measured using task accuracy or semantic similarity metrics that explicitly tolerate paraphrase and abstraction, such as BERTScore \cite{zhang2019bertscore}, BLEURT \cite{sellam2020bleurt}, and COMET \cite{rei2020comet}. 

Conversational speech poses a different challenge. Disfluency removal is a deletion-constrained task requiring the fluent output to preserve the original token sequence except for annotated disfluencies. Unlike semantic benchmarks that permit paraphrasing, this setting demands strict structural fidelity, where additional deletions or rewrites are errors. Robustness must therefore be measured by agreement with the transcript’s structural constraints rather than semantic equivalence.

\section{Robustness to Conversational Structure: DRES as a Factorized Structural Evaluation Framework}
\label{Robustness to Conversational Structure}

We formalize robustness to conversational structure as fidelity to a uniquely specified deletion-only transformation over conversational transcripts, as shown in \textbf{\autoref{fig:system}}. 

Let $x$ denote a tokenized transcript containing disfluencies, and let $m \in \{0,1\}^n$ denote its gold disfluency mask. Because disfluency removal is deletion-constrained and uniquely determined under gold annotation, robustness can be defined in terms of agreement with this mask. DRES operationalizes this definition by isolating language-level deletion behavior from acoustic variability and evaluating agreement with $m$ directly.

\renewcommand{\arraystretch}{1.0}
\begin{table*}
    \centering
    \small
    \resizebox{0.99\textwidth}{!}{%
    \begin{tabular}{
      @{}
      >{}p{1.9cm} 
      >{}p{1.0cm} 
      >{}p{1.0cm} 
      >{}p{1.2cm} 
      >{}p{4.1cm} 
      >{}p{2.1cm} 
      >{}p{2.1cm} 
      >{\small}p{5.2cm} 
      @{}
    }
        \toprule
             {\normalfont \rmfamily \textit{LLM}} & 
            {\normalfont \rmfamily \textit{Citation}} & 
            {\normalfont \rmfamily \textit{Open-Source}}& 
            {\normalfont \rmfamily \textit{Instruct}}& 
            {\normalfont \rmfamily \textit{Sizes}} & 
            {\normalfont \rmfamily \textit{Architecture}} & 
            {\normalfont \rmfamily \textit{Context Length}} & 
            {\normalfont \rmfamily \textit{Features}}
        \\
        \midrule
        gpt-4o & \cite{achiam2023gpt} & \xmark & \cmark & Nx200B* & MoE* & 128k & Multimodal, Instruct
        \\
        gpt-4o-mini & \cite{achiam2023gpt} & \xmark & \cmark & Nx8B* & MoE* & 128k & Multimodal, Instruct
        \\
        o4-mini & \cite{achiam2023gpt} & \xmark & \cmark & 100B & Dense & 200k & Multimodal, Instruct, Reasoning
        \\
        Llama-3.1 & \cite{llama3_2024} & \cmark & \cmark & 8B & Dense & 128k & Instruct
        \\
        Llama-3.2 & \cite{llama3_2024} & \cmark & \cmark & 1B,3B & Dense & 128k & Instruct
        \\
        Llama-3.3 & \cite{llama3_2024} & \cmark & \cmark & 70B & Dense & 128k & Multimodal, Instruct
        \\
        MobileLLM & \cite{liu2024mobilellmoptimizingsubbillionparameter} & \cmark & \xmark & 125M, 350M, 600M, 1B & Dense & 2048+ & Small for edge devices
        \\
        Qwen3 & \cite{qwen3technicalreport} & \cmark & \cmark & 0.6B,1.7B, 4B, 8B & Dense + MoE & 32,768+ &  Instruct
        \\
        Phi-4-mini & \cite{microsoft2025phi4minitechnicalreportcompact} & \cmark & \cmark & 3.8B & Dense & 128k & Instruct, Reasoning
        \\
        \bottomrule
    \end{tabular}
    }
    \caption{\textbf{Backbone LLMs Evaluated}: * indicates rumored sizes \cite{zeff2024openai, epoch2024frontierlanguagemodelshavebecomemuchsmaller, abacha2025medecbenchmarkmedicalerror}. o4-mini is available in high/medium reasoning variants, we extrapolate the size from rumored o1-mini sizes. While some include multimodal capabilities, they are primarily text-based language models.}
    \label{tab:model-comparison}
\end{table*}

\subsection{SpeechLLMs as Composed Systems}
\label{SpeechLLMs as Composed Systems}

To operationalize structural robustness, we first characterize how deletion behavior arises within a SpeechLLM pipeline. A SpeechLLM composes an acoustic encoder $A$ with a language backbone $L_\theta$ \cite{yang-etal-2025-large-language, cui2025recent, arora2025landscape, Peng_2026}:
\[
A: W \rightarrow X, 
\qquad
L_\theta: X \rightarrow X,
\]
yielding the end-to-end mapping
\[
y = L_\theta(A(w)).
\]

Deletion behavior therefore arises from two operators: acoustic suppression in $A$ and structural editing in $L_\theta$.

\subsection{Acoustic Suppression and Editing Bias}
\label{Acoustic Suppression and Editing Bias}

This decomposition highlights a key measurement challenge: deletions in the final transcript may originate either from the acoustic encoder or from the language model’s editing policy. Without separating these effects, structural failures cannot be attributed to the backbone model itself.

Modern ASR systems under-transcribe certain disfluencies, especially short interjections \cite{mujtaba2024lost, teleki24_interspeech}. We model acoustic omission as a deletion mask
\[
S_A : x \mapsto \hat{x}, 
\qquad 
\hat{x} = (x_i : s_i = 0), \quad s \in \{0,1\}^n.
\]

Observed end-to-end deletions reflect the composition
\[
x \xrightarrow{S_A} \hat{x} \xrightarrow{D_\theta} y.
\]

Without controlling $S_A$, deletions cannot be attributed to acoustic omission versus backbone editing. Even when $S_A = 0$ (perfect transcription), the backbone model may fall into one of the four editing policies; thus systematic over- or under-deletion arises from the backbone model $D_\theta$ itself, reflecting training objectives that favor abstraction or compression rather than structural fidelity.
DRES fixes $S_A = 0$ and evaluates
\[
R_\theta(x) = R(D_\theta(x), m),
\]
thereby isolating backbone-level editing policies.

\subsection{Robustness and Editing Policy Definitions}
\label{Robustness and Editing Policy Definitions}

Having isolated backbone behavior, we can now define robustness directly in terms of the model’s deletion decisions. Let $x$ be a transcript, $m \in \{0,1\}^n$ its gold deletion mask, and $D_\theta(x) \in \{0,1\}^n$ the model-implied deletion mask recovered via alignment \cite{teleki25_zscores}. We define:

\[
TP_\theta(x) := \sum_i \mathbbm{1}[m_i=1 \land D_{\theta,i}(x)=1],
\]
\[
O_\theta(x) := \sum_i \mathbbm{1}[m_i=0 \land D_{\theta,i}(x)=1],
\]
\[
U_\theta(x) := \sum_i \mathbbm{1}[m_i=1 \land D_{\theta,i}(x)=0].
\]

Here $TP_\theta(x)$ denotes true positive deletions (i.e., the token should be deleted in reference to the gold transcript and it was deleted by $L_{\theta}$), $O_\theta(x)$ denotes over-deletions and $U_\theta(x)$ under-deletions. 
Word-level precision and recall ($\mathcal{E}$-Scores) can be written directly as

\[
\mathcal{E}_P(x;\theta)
=
\frac{TP_\theta(x)}{TP_\theta(x)+O_\theta(x)},
\quad
\mathcal{E}_R(x;\theta)
=
\frac{TP_\theta(x)}{TP_\theta(x)+U_\theta(x)}.
\]

\noindent \textbf{Definition 1: Robustness via Editing Policies.} Thus robustness is defined in terms of $\mathcal{E}$-Scores as:
\[
R_\theta(x) = R\big(D_\theta(x), m\big) \\
= \{\mathcal{E}_P,\mathcal{E}_R,\mathcal{E}_F\}\big(D_\theta(x), m\big) 
\]
and it is fully determined by the pair $(O_\theta(x), U_\theta(x))$. This definition quantifies how robust the backbone model $L_{\theta}$ is to over-deletion and under-deletion.

\medskip
\noindent \textbf{Editing Policies.} These quantities naturally induce a geometric interpretation of model behavior in precision–recall space. Different relative magnitudes of over- and under-deletions correspond to distinct editing regimes, which we refer to as editing policies. There are four editing policies in $(\mathcal{E}_P,\mathcal{E}_R)$-space:

\[
\begin{array}{ll}
\textcolor{conservative}{\textbf{Under-Deletion}}
& U_\theta(x) \gg O_\theta(x) 
\Rightarrow (\mathcal{E}_P \uparrow, \mathcal{E}_R \downarrow) \\

\textcolor{aggressive}{\textbf{Over-Deletion}}
& O_\theta(x) \gg U_\theta(x) 
\Rightarrow (\mathcal{E}_P \downarrow, \mathcal{E}_R \uparrow) \\

\textcolor{balanced}{\textbf{Balanced}}
& O_\theta(x), U_\theta(x) \text{ both small} 
\Rightarrow (\mathcal{E}_P \uparrow, \mathcal{E}_R \uparrow) \\

\textcolor{poor}{\textbf{Poor}}
& O_\theta(x), U_\theta(x) \text{ both large} 
\Rightarrow (\mathcal{E}_P \downarrow, \mathcal{E}_R \downarrow)
\end{array}
\]

\begin{figure*}
    \centering
    \includegraphics[width=0.80\linewidth]{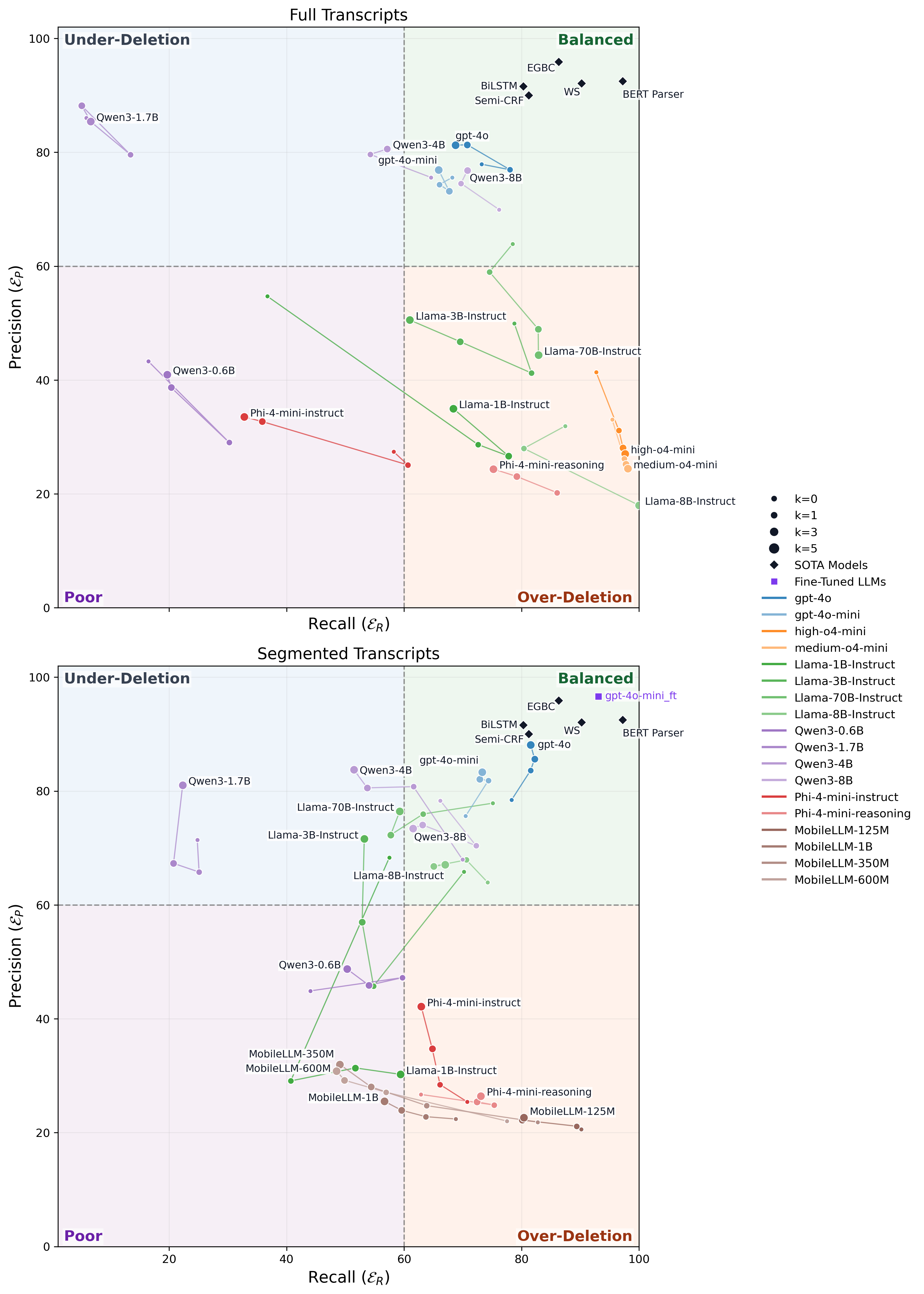}
\caption{\footnotesize Precision--recall trade-offs across models under full (top) and segmented (bottom) transcripts. Each point corresponds to a model evaluated at varying in-context learning levels ($k=\{0,1,3,5\}$). Shaded quadrants reveal four \textbf{editing policies}: \\
\textcolor{conservative}{\textbf{Under-Deletion} ($\mathcal{E}_P \uparrow$, $\mathcal{E}_R \downarrow$)} occurs when models fail to recognize conversational structure and leave many disfluencies untouched. \\
\textcolor{aggressive}{\textbf{Over-Deletion} ($\mathcal{E}_P \downarrow$, $\mathcal{E}_R \uparrow$)} reflects a rewriting bias: models treat the task as paraphrasing and delete fluent words. \\
\textcolor{balanced}{\textbf{Balanced} ($\mathcal{E}_P \uparrow$, $\mathcal{E}_R \uparrow$)} represents the desired behavior, combining accurate disfluency identification with preservation of fluent content. \\
\textcolor{poor}{\textbf{Poor} ($\mathcal{E}_P \downarrow$, $\mathcal{E}_R \downarrow$)} inherits the worst of both behaviors, missing disfluencies while also deleting fluent tokens. \\
\underline{Proprietary models} cluster in the \textcolor{balanced}{Balanced} region, while \underline{reasoning models} cluster in the \textcolor{aggressive}{Over-Deletion} region; \underline{small models} more frequently occupy the \textcolor{aggressive}{Over-Deletion} or \textcolor{poor}{Poor} regimes.
Qualitative structure remains consistent for thresholds in the range $0.55$--$0.70$; quantitative clustering analysis appears in \S\ref{Empirical Evidence of Policy-Level Robustness Patterns} ($\rhd$ Findings 1, 3, 4, 5).}
    \label{fig:main}
\end{figure*}

\begin{figure}
    \centering
    \vspace{-10pt}
    \includegraphics[width=0.45\linewidth]{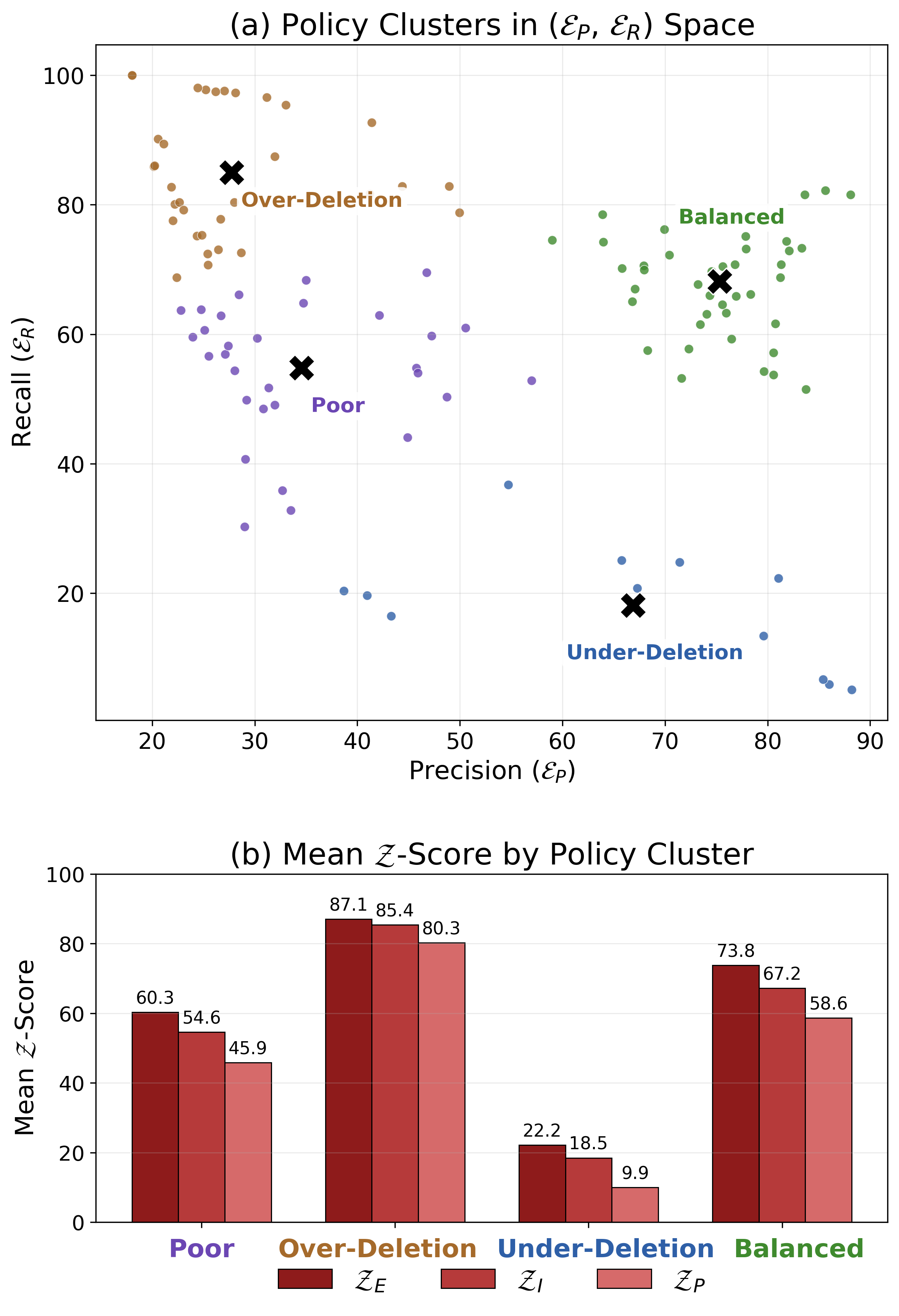}
    \caption{Policy structure in robustness space. 
(a) Clustering in $(\mathcal{E}_P,\mathcal{E}_R)$ space recovers groups that align with the quadrant regimes in Figure~\ref{fig:main}, indicating that \textbf{editing policies emerge as geometric structure in robustness space}. Black markers denote cluster centers ($\rhd$ Finding 1). 
(b) Mean category-level $\mathcal{Z}$ scores for each policy cluster ($\mathcal{Z}_E$, $\mathcal{Z}_I$, $\mathcal{Z}_P$) show consistent behavioral differences across disfluency types ($\rhd$ Finding 2).}
    \label{fig:clustering}
\end{figure}

We show these policies in \autoref{fig:main}. 
As discussed in \S \ref{Quantitative Validation of Editing Policies}, clustering modeling in this space recovers groups that align with these regions (\autoref{fig:clustering}), indicating that editing policies emerge as natural geometric structure in robustness space.

\medskip
\noindent \textbf{Definition 2: Robustness via Disfluency Categories.} We also define robustness in terms of \textcolor{zcolor}{\textbf{category-specific}} $\mathcal{Z}$-Scores \cite{teleki25_zscores}:
\[
R_\theta(x) = R\big(D_\theta(x), m\big) \\
= \{\mathcal{Z}_E,\mathcal{Z}_I,\mathcal{Z}_P\}\big(D_\theta(x), m\big) 
\]

Which quantify how robust the backbone model $L_{\theta}$ is to the three types of disfluencies in the Shriberg definition \cite{Shriberg.1994}: \textcolor{zcolor}{\textbf{EDITED}}, \textcolor{zcolor}{\textbf{INTJ}}, and \textcolor{zcolor}{\textbf{PRN}}, and all other parts-of-speech are considered `fluent,' as shown in \autoref{fig:system}.

\begin{figure}
    \centering
    \includegraphics[width=0.48\linewidth]{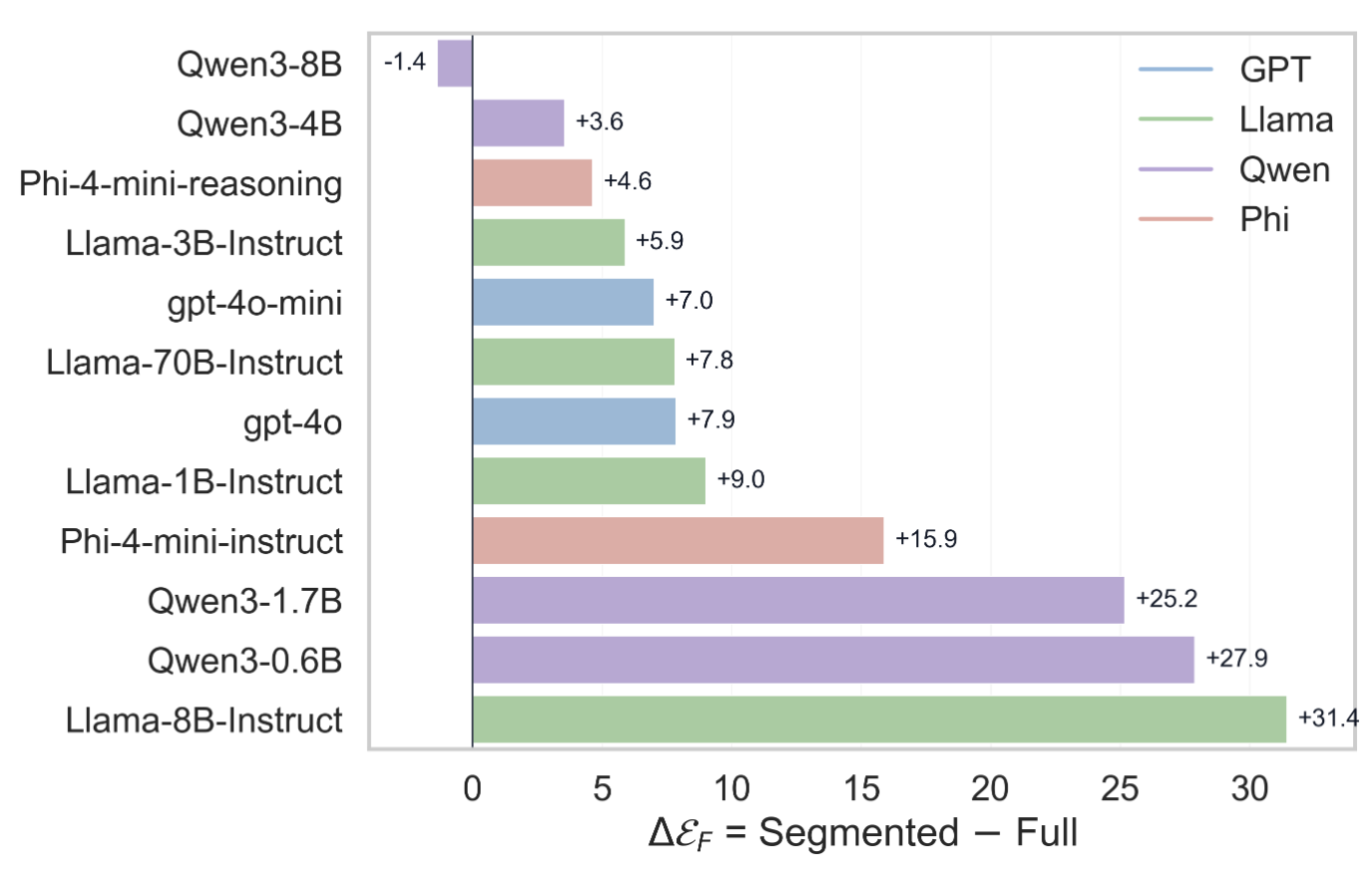}
    \caption{\textbf{Segmentation improves robustness to conversational structure}. Segmenting long conversational transcripts consistently improves structural fidelity across models. Performance gains in $\Delta\mathcal{E}_F$ indicate that robustness failures largely arise from long-context instability rather than knowledge limitations. Averaged across $k$ ($\rhd$ Finding 3).}
    \label{fig:segmentation}
\end{figure}

\subsection{Switchboard Dataset}
\label{Switchboard Dataset}

DRES is built on the Switchboard Treebank \cite{mitchell1999treebank, godfrey1992switchboard}, which provides gold parse trees with paired fluent and disfluent realizations. Disfluencies are defined using the Shriberg scheme \cite{Shriberg.1994}: \textcolor{zcolor}{\textbf{EDITED}}, \textcolor{zcolor}{\textbf{INTJ}}, and \textcolor{zcolor}{\textbf{PRN}} are labeled disfluent; all other parts-of-speech are fluent \cite{jamshid-lou-johnson-2020-improving,Charniak.2001}. Fluent and disfluent transcripts are constructed recursively from gold trees \cite{Charniak.2001}, retaining partial words and punctuation to match modern ASR outputs \cite{johnson-charniak-2004-tag}. Because Switchboard transcripts have been manually produced and, importantly, they have been iteratively corrected \cite{lickley1996not}, they provide reliable structural targets. Gold transcripts are essential, as ASR systems systematically under-transcribe disfluencies \cite{mujtaba2024lost, teleki24_interspeech}.

\begin{figure}
    \centering
    \includegraphics[width=0.65\linewidth]{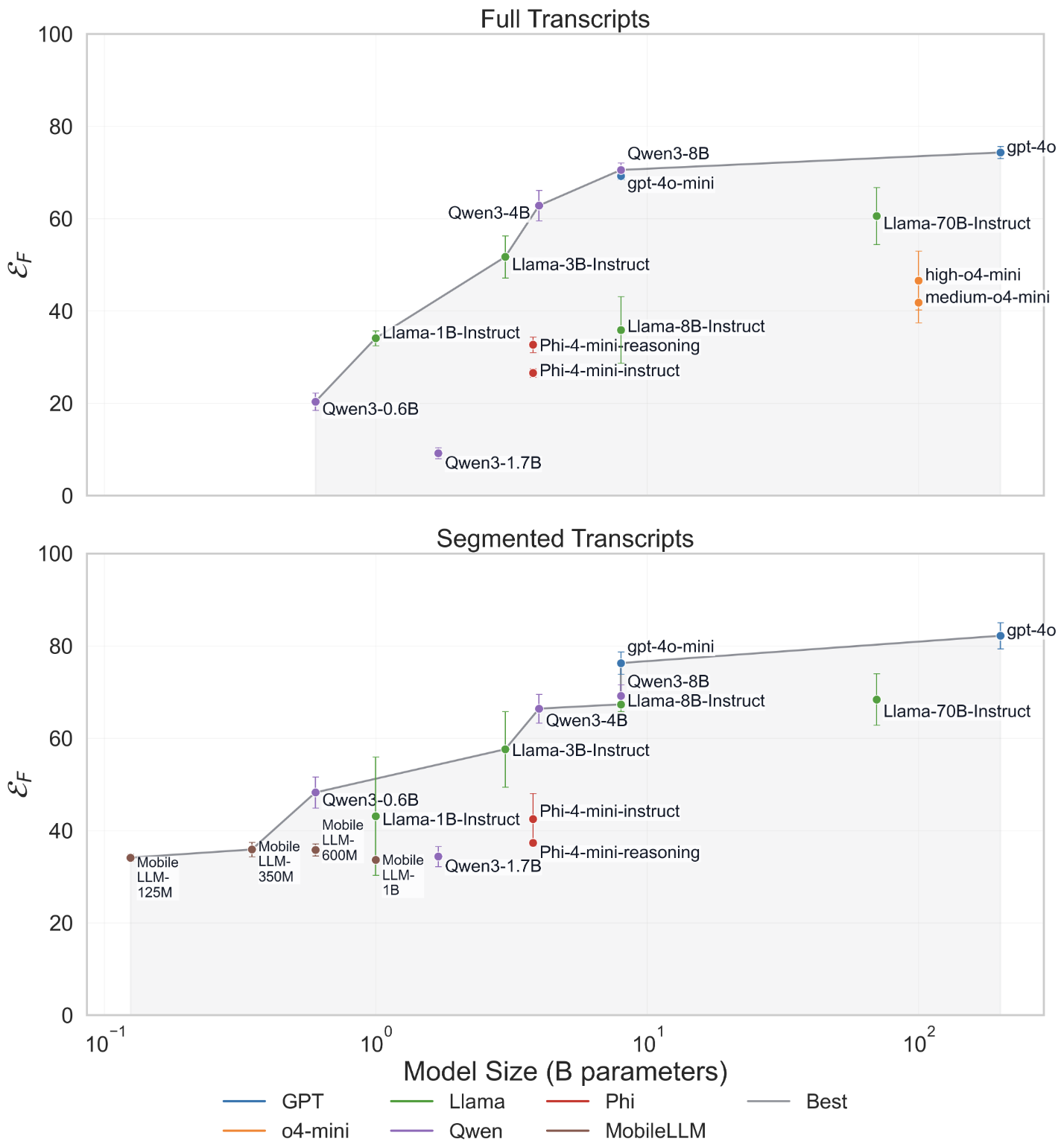}
    \caption{\textbf{Scale improves performance, but does not change editing policy.} Larger models within each family achieve higher $\mathcal{E}_F$ but reasoning-oriented variants consistently underperform relative to the best performance curve. This shows that scale improves execution of a policy but does not change the underlying editing behavior.($\rhd$ Finding 4)}
    \label{fig:model-scale}
\end{figure}

\subsection{Evaluation Protocol}
\label{Evaluation Protocol}

\textbf{Deletion Alignment.}
Token-level deletion decisions are recovered by alignment, using the methodology of \cite{teleki25_zscores}. 

\smallskip
\noindent \textbf{Context Conditions.}
Models are evaluated under: full transcripts, and segmented transcripts (approximately 4 sentences each). Segmentation shortens context while preserving local structure, enabling separation of intrinsic editing policy from long-context instability.

\smallskip
\noindent \textbf{Fine-Tuning and Generalization.}
To measure specialization effects, models are evaluated before and after adaptation on GSM8K \cite{gsm8k}, MMLU \cite{hendrycks2020measuring}, and CoQA \cite{reddy2019coqa}.

\smallskip
\noindent \textbf{In-Context Learning ($k$).}
We evaluate models under in-context learning \cite{brown2020language}, where the model is conditioned on a small number of task demonstrations provided directly in the prompt rather than through parameter updates. Example pairs are drawn from the development portion of the dataset and formatted as disfluent--–fluent pairs, illustrating the deletion-only transformation the model should perform. The ordering of demonstrations is fixed across models to ensure consistent evaluation. Here $k$ denotes the number of demonstration pairs included in the prompt prior to the evaluation instance. We evaluate models with $k \in \{0,1,3,5\}$, shown in \autoref{fig:main}.

\smallskip
\noindent \textbf{Comparison to Previous SOTA.}
Prior work on conversational disfluency primarily treats the problem as supervised span detection using sequence labeling models such as BiLSTM and Semi-CRF \cite{zayats2016disfluency}, EGBC \cite{bach2019noisy}, weight-sharing approaches \cite{wang2018semi}, and BERT-based parsers \cite{jamshid-lou-johnson-2020-improving} (performance shown in \autoref{fig:main}). In contrast, SpeechLLMs rely on generative LLM backbones that perform disfluency removal through open-ended editing rather than explicit span prediction. DRES evaluates these generative repairs under the same structural criteria used in prior work, enabling direct comparison between sequence-labeling and generative approaches.

\subsection{Model Axes}
\label{Model Axes}

We evaluate proprietary and open-source LLMs (\autoref{tab:model-comparison}) spanning: \textbf{model size} (125M to frontier scale); \textbf{architecture} (dense vs.\ mixture-of-experts); \textbf{instruction-tuned vs.\ base vs.\ reasoning} variants; \textbf{context window length}. This controlled design allows comparison of how scale, objective, architecture, and context management shape deletion-constrained repair.

\section{Empirical Evidence of Policy-Level Robustness Patterns}
\label{Empirical Evidence of Policy-Level Robustness Patterns}

\autoref{fig:main} plots model performance in the precision–recall $(E_P, E_R)$ space under full and segmented transcript conditions. 

\subsection{Quantitative Validation of Editing Policies}
\label{Quantitative Validation of Editing Policies}

To verify that the \textit{editing policies} described in \S \ref{Robustness and Editing Policy Definitions} (\textcolor{conservative}{\textbf{Under-Deletion}}, 
\textcolor{aggressive}{\textbf{Over-Deletion}}, 
\textcolor{balanced}{\textbf{Balanced}}, and
\textcolor{poor}{\textbf{Poor}}) reflect intrinsic structure, we analyze model behavior in the $(E_P, E_R)$ space defined in \S\ref{Robustness and Editing Policy Definitions}. 
We perform a $k$-means clustering analysis in the $(E_P, E_R)$ plane. To select the number of clusters, we compute the silhouette score ($S$) across $k \in \{2,\dots,7\}$ (Figure~\ref{fig:clustering}). The score peaks at $k=4$ and declines for larger $k$. The silhouette score ($S=0.54$) and Davies-Bouldin Index ($DBI=0.664$) indicate moderate-to-strong cluster separation and internal cohesion, confirming that four clusters provide the best separation in the $(E_P,E_R)$ space. 

A permutation test with 200 randomizations produced no silhouette scores exceeding the observed value ($p \approx 0.005$), indicating that the cluster structure is unlikely under a random $(E_P,E_R)$ pairing. This indicates that the editing polices indeed correspond to natural groupings in this space.

We find that cluster assignments are largely stable within model families. Across prompting levels, models retain their policy classification in 70–100\% of cases, indicating that editing behavior is a backbone-intrinsic property. However, stability differs based on whether we compare performance over full transcripts or segmented transcripts: while GPT models exhibit perfect policy stability across both settings ($p=1.00$), the Llama family shows significant volatility ($p=0.25$), with Phi ($p=0.75$) and Qwen ($p=0.79$) maintaining moderate consistency in their cluster assignments across both settings.

\subsection{$\rhd$ Finding 1: Editing policies align with training objectives (Figures \ref{fig:main}, \ref{fig:clustering}).}
\label{Finding 1}

Models consistently occupy distinct precision–recall regimes corresponding to conservative, aggressive, balanced, and poor editing behavior. GPT models cluster in the balanced region across prompting conditions, while reasoning-oriented models (e.g., o4-mini and Phi-4 reasoning variants) systematically shift toward aggressive over-deletion. Smaller or base models frequently adopt conservative under-deletion policies, preserving fluent tokens while failing to remove many disfluencies.

These patterns remain stable across prompting levels and scale within families, indicating that editing behavior is primarily determined by training objectives rather than parameter count.

\begin{figure*}
    \centering
    \includegraphics[width=0.99\linewidth]{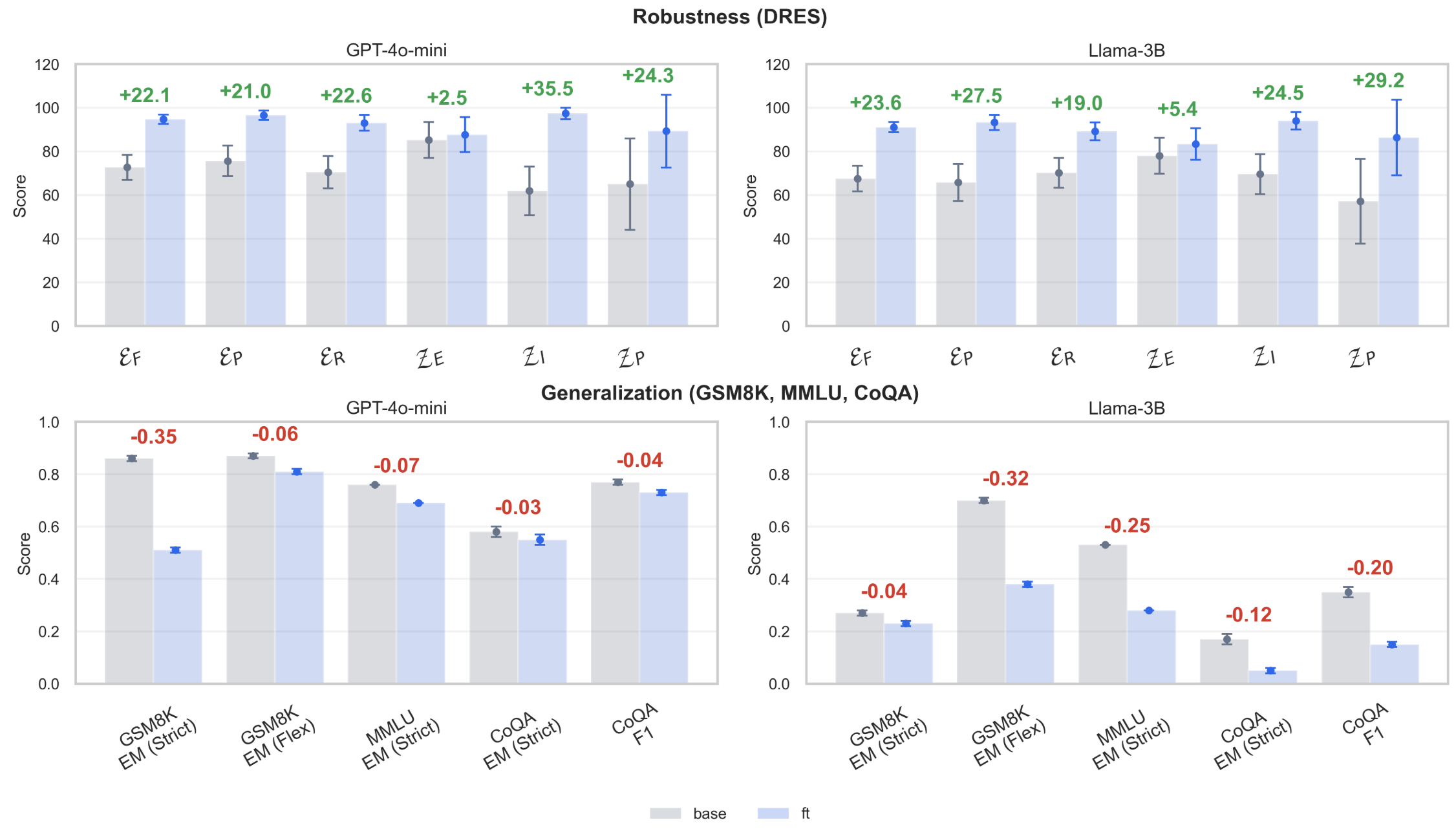}
    \caption{\textbf{Fine-Tuning Results} ($\rhd$ Finding 5).}
    \label{fig:fine-tuning}
\end{figure*}

\begin{table*}[t]
\centering
\small
\begin{tabular}{lcccc}
\toprule
\textbf{Model} & \textbf{$\Delta$DRES (95\% CI)} & \textbf{$\Delta$MMLU (95\% CI)} & \textbf{$\Delta$GSM8K (95\% CI)} & \textbf{$\Delta$CoQA (95\% CI)} \\
\midrule
GPT-4o-mini 
& $\posval\posmain{22.05}\ [\ \posval 9.83,\ \posval 34.10\ ]$
& $\negval\negmain{0.07}\ [\ \negval 0.07,\ \negval 0.07\ ]$
& $\negval\negmain{0.35}\ [\ \negval 0.38,\ \negval 0.32\ ]$
& $\negval\negmain{0.03}\ [\ \negval 0.09,\ \posval 0.02\ ]$ \\

Llama-3B    
& $\posval\posmain{23.46}\ [\ \posval 10.90,\ \posval 35.91\ ]$
& $\negval\negmain{0.25}\ [\ \negval 0.25,\ \negval 0.25\ ]$
& $\negval\negmain{0.04}\ [\ \negval 0.07,\ \negval 0.01\ ]$
& $\negval\negmain{0.12}\ [\ \negval 0.16,\ \negval 0.08\ ]$ \\
\bottomrule
\end{tabular}
\caption{Paired Base $\rightarrow$ Fine-Tuned Performance Changes.
$\Delta$DRES reflects improvement in structural fidelity (segmented condition, $\mathcal{E}_F$).
Negative $\Delta$ values indicate degradation on generalization benchmarks.
Confidence intervals are 95\%, ($\rhd$ Finding 5).}
\label{tab:ft_tradeoff}
\end{table*}

\subsection{$\rhd$ Finding 2: Models handle overt repairs well (\textcolor{zcolor}{EDITED}) but struggle with short conversational markers (\textcolor{zcolor}{INTJ}, \textcolor{zcolor}{PRN}) (\autoref{fig:clustering}b).}
\label{Finding 2}

Category-level $\mathcal{Z}$-scores reveal systematic differences across disfluency types (\autoref{fig:clustering}b). The models perform well on \textcolor{zcolor}{\textbf{EDITED}} structures, commonly studied in prior disfluency detection work \cite{jamshid-lou-johnson-2020-improving,johnson-charniak-2004-tag}.

In contrast, performance drops substantially on \textcolor{zcolor}{\textbf{INTJ}} (interjections such as ``uh'' and ``um'') and \textcolor{zcolor}{\textbf{PRN}} (parenthetical insertions such as ``you know'' and ``I mean''). These markers are short, frequent, and embedded within otherwise fluent structure. Notably, this result contrasts with prior work suggesting that these categories are among the easiest to detect and remove \cite{jamshid-lou-johnson-2020-improving,johnson-charniak-2004-tag}. This discrepancy indicates that generative models encounter \textit{different failure modes} than traditional sequence-tagging approaches (\S \ref{Disfluency Detection and Removal}).

The pattern largely persists across model families and scales, suggesting a training distribution mismatch: conversational markers common in spontaneous speech appear underrepresented in pretraining corpora.

\subsection{$\rhd$ Finding 3: Long transcripts expose context instability rather than knowledge gaps (Figures \ref{fig:main}, \ref{fig:segmentation}).}
\label{Finding 3}

Performance differences between full transcripts and segmented inputs reveal a context-management failure mode.\footnote{This pattern is consistent with prior observations that long-context language models often handle information appearing in the middle of a sequence less reliably than information near its boundaries \cite{liu2024lost}.} Under full transcripts, several models exhibit unstable precision–recall behavior, frequently collapsing into \textcolor{aggressive}{\textbf{over-deletion}} regimes.
Segmenting transcripts into shorter contexts consistently shifts models toward the \textcolor{balanced}{\textbf{balanced}} region and reduces variance in performance. This improvement occurs even for models with large context windows, indicating that instability arises from context handling rather than insufficient capacity.
Thus, robustness failures in conversational speech partly reflect architectural sensitivity to long-context structure.

\subsection{$\rhd$ Finding 4: Scale generally improves performance within model families but does not change the editing policy (Figures \ref{fig:main}, \ref{fig:model-scale}).}
\label{Finding 4}

Structural performance generally improves with model size (Figures \ref{fig:main}, \ref{fig:model-scale}). Within each model family, larger variants consistently achieve higher $\mathcal{E}_F$, forming an improvement curve as parameter count increases.

However, \textit{scale does not alter the underlying editing policy}. Models tend to remain within the same precision–recall policy as they grow. Conservative models become more precise but continue to \textcolor{conservative}{\textbf{under-delete}} disfluencies, while aggressive models improve recall yet still \textcolor{aggressive}{\textbf{over-delete}} fluent content.

This pattern indicates that scale primarily refines performance along a family-specific trajectory rather than changing the qualitative editing behavior itself. The editing policy therefore appears to be determined by training objectives and alignment choices, while scale controls how well the model executes that policy.

\subsection{$\rhd$ Finding 5: Fine-tuning introduces a robustness--generalization trade-off, improving structural fidelity but reducing generalization (Figures \ref{fig:main}, \ref{fig:fine-tuning}, \autoref{tab:ft_tradeoff}).}
\label{Finding 5}

Fine-tuning substantially improves structural fidelity on the disfluency removal task. As shown in Figures \ref{fig:main} and \ref{fig:fine-tuning}, both models move close to the upper bound of the DRES metrics after adaptation, with $\mathcal{E}_F$ rising from the mid–70s to the mid–90s for GPT-4o-mini and from the high–60s to above 90 for Llama-3B. Table~\ref{tab:ft_tradeoff} summarizes these gains as improvements of more than \textbf{+22 points} in overall DRES performance for both models.

This improvement comes at a cost. Across GSM8K, MMLU, and CoQA, both models exhibit lower post–fine-tuning scores (Figure~\ref{fig:fine-tuning}), with the largest decline appearing on \textit{reasoning-heavy tasks} such as GSM8K. Recent studies have also shown this same phenomenon: fine-tuning LLMs on domain-specific datasets can substantially impair their generalization capabilities  \cite{chu2025sft, chen2025sft, huan2025does, lin2025sft}.

Together, these results reveal a robustness–generalization trade-off: fine-tuning aligns models with deletion-constrained structural repair, but this specialization narrows their broader reasoning and knowledge capabilities.

\begin{table*}[ht]
\centering

\renewcommand{\arraystretch}{1.1}
\begin{tabular}{@{}llp{13cm}@{}}
\toprule
\textbf{Category} &  & \textbf{Guideline for Deployment and Development} \\ \midrule
\textbf{Deployment} & R1 & Use small, \textcolor{conservative}{\textbf{under-deletion}} models (e.g., Qwen3-1.7B) for edge-based, real-time intent preservation \\
 & R2 & Prioritize segmented input over full transcripts to ensure stability, regardless of context window size \\
 & R3 & Leverage 8B-class models with few-shot prompting for a reasonable latency-accuracy trade-off \\ \midrule
\textbf{Objective} & R4 & Avoid reasoning-oriented models for literal repair to prevent \textcolor{aggressive}{\textbf{over-deletion}} \\
 & R5 & In command-driven interfaces, select models with a \textcolor{conservative}{\textbf{under-deletion}} policy to protect fluent nouns/verbs \\
 & R6 & Account for stable family-wide \textbf{editing policies} when scaling; larger models retain similar error biases \\ \midrule
\textbf{Training} & R7 & Use domain-specific fine-tuning for archival-grade transcription where SOTA precision is required \\
 & R8 & Monitor the generalization tax; evaluate MMLU/GSM8K scores when fine-tuning for speech tasks \\
 & R9 & Utilize DRES as a diagnostic structural probe to audit new LLMs before integration into SpeechLLMs \\ \bottomrule
\end{tabular}
\caption{Practical Recommendations (R1--R9) for Conversational Robustness in Speech Pipelines}
\label{tab:recommendations}
\end{table*}

\subsection{Practical Recommendations (\autoref{tab:recommendations})}
\label{Practical Recommendations}

The empirical results from our DRES evaluation translate into several deployment guidelines for SpeechLLM pipelines (Table~\ref{tab:recommendations}). Segmenting transcripts improves robustness even for long-context models, while small high-precision models suit edge settings and mid-scale models with light prompting provide strong latency–accuracy trade-offs (R1–R3). Editing behavior is driven primarily by training objectives rather than scale -- reasoning-oriented models often over-delete fluent content—while fine-tuning improves structural fidelity but introduces a measurable generalization cost, motivating structural diagnostics such as DRES before deployment (R4–R9).

\section{Limitations}
This study focuses on structural identifiability rather than full end-to-end deployment behavior.  DRES fixes acoustic suppression and measures deletion decisions against gold masks, enabling structural editing policies to be analyzed independently of ASR errors, though real systems combine both effects. Experiments are conducted on English Switchboard transcripts; while the deletion-constrained framework is language-agnostic, conversational repair patterns may vary across languages and remain to be validated. Finally, the deletion-only task captures minimal structural fidelity but does not model broader normalization behaviors.

\section{Future Directions} 
Beyond typical conversational disfluency, clinical speech conditions such as aphasia, dysarthria, and other language production disorders introduce different structural phenomena \cite{macwhinney2011aphasiabank, mei2025addressingpitfallsauditingpractices}. Future work should develop condition-aware structural auditing frameworks and incorporate domain-specific safeguards when deploying SpeechLLMs in healthcare environments.

\section{Conclusion}
\label{Conclusion}

With DRES, we show that conversational speech reveals systematic, objective-induced limits in current LLMs. Robustness to conversational structure does not improve monotonically with scale. Instead, models adopt stable editing biases shaped by training objectives and context handling.
Deletion-only structural tasks expose these tendencies with minimal ambiguity. Aggressive semantic abstraction conflicts with constrained repair; conservative preservation fails to remove true disfluencies. Fine-tuning can improve local structural fidelity, but at measurable cost to broader generalization.
Structural diagnostics such as DRES therefore provide a complementary evaluation axis for SpeechLLM development.

\section{Generative AI Use Disclosure}
AI tools supported parts of the coding process and helped structure and clarify written content. All generated material was carefully reviewed, validated, and revised by the authors.

\newpage

\bibliographystyle{IEEEtran}
\bibliography{mybib}

\end{document}